\documentclass[journal,transmag]{IEEEtran}

\usepackage{graphicx}
\usepackage{subcaption}
\usepackage{float}
\usepackage{tabularx}
\usepackage{rotating}
\usepackage{multirow}
\usepackage{cite}
\usepackage{amsmath}
\usepackage{amssymb}
\usepackage{nameref}
\usepackage{flushend}

\begin{document}

	\title{Single neuron-based neural networks are as efficient as dense deep neural networks in binary and multi-class recognition problems}
	
	\author{Yassin Khalifa, Justin Hawks, and Ervin Sejdi\'{c}, \IEEEmembership{Senior Member,~IEEE,}}
	
	\maketitle

	\begin{abstract}
		Recent advances in neuroscience have revealed many principles about neural processing. In particular, many biological systems were found to reconfigure/recruit single neurons to generate multiple kinds of decisions. Such findings have the potential to advance our understanding of the design and optimization process of artificial neural networks. Previous work demonstrated that dense neural networks are needed to shape complex decision surfaces required for AI-level recognition tasks. We investigate the ability to model high dimensional recognition problems using single or several neurons networks that are relatively easier to train. By employing three datasets, we test the use of a population of single neuron networks in performing multi-class recognition tasks. Surprisingly, we find that sparse networks can be as efficient as dense networks in both binary and multi-class tasks. Moreover, single neuron networks demonstrate superior performance in binary classification scheme and competing results when combined for multi-class recognition.
	\end{abstract}

	{
		\renewcommand{\thefootnote}{}%
		\footnotetext[1]{Research reported in this publication was supported by the National Science Foundation under the CAREER Award Number 1652203. The content is solely the responsibility of the authors and does not necessarily represent the official views of the National Science Foundation.}
		\footnotetext[2]{Yassin Khalifa, Justin Hawks and Ervin Sejdi\'{c} are with the Department of Electrical and Computer Engineering, Swanson School of Engineering, University of Pittsburgh, Pittsburgh, PA, 15261, USA (e-mail: esejdic@ieee.org).}
	}
	\IEEEpeerreviewmaketitle
	
	\section{Introduction}
	\label{S:Introduction}
	Over the last several years, machine learning techniques, particularly neuromorphic computing architectures, have been used to mimic the challenging pattern recognition abilities of biological systems \cite{lecun2015deep}. In fact, it could be argued that neural networks based systems are learning feature hierarchies automatically so that they are replacing many learning models that depend on hand-designed heuristics \cite{lecun2015deep,lecunCNN,replearnBengio}. This is achieved through multiple levels of representations generated by successive non-linear modules, each of which transforms low level representations into more abstract high level representations \cite{lecun2015deep,XavBenjio_2010, Vincent_2008,chen2017convolutional,ma2018using}. Theoretical results have shown that deep architectures may be needed to efficiently model complex functions and represent high levels of abstraction required for challenging recognition and AI tasks \cite{bengiolecun2007,Bengio:2009,Vincent_2008,zhu2018image}.\par
	
	Many questions come up when choosing an architecture to address a practical problem, including how many layers we should use and how large each layer is supposed to be \cite{Goodfellow-et-al-2016,NIPS2014_5484}. Network capacity increases with both number and size of hidden layers due to the fact that neurons collaborate to express complex functions and achieve better generalization \cite{Goodfellow-et-al-2016}. However, high capacity models may fit also dataset outliers and lead to overfitting. Although, one can always use shallow networks to avoid overfitting, but sometimes data complexity necessitates using deep architectures \cite{Goodfellow-et-al-2016,losssurfaces}. Multiple solutions have been suggested though to avoid overfitting in deep networks including dropout \cite{srivastava2014dropout}, regularization, and noisy inputs \cite{Goodfellow-et-al-2016,vangorp, losssurfaces}.\par
	
	In practice, overfitting control is preferred over network size tuning as it was found that smaller networks are hard to train using local optimizers like gradient descent \cite{losssurfaces,LeCun1998,saxe2013exact,dauphin2014identifying}. Obviously, smaller networks will have fewer local solutions with easy convergence, however, most of these solutions are unreliable (high loss) especially when using batch optimization \cite{losssurfaces}. This also makes small sized networks vulnerable to huge loss variance influenced by random initializations \cite{Goodfellow-et-al-2016}. It is hard to prove this mathematically due to the poor understanding of loss function and its shape for neural networks \cite{losssurfaces}. On the other hand, simplifying the problem through conquering into smaller problems or multiple recognition stages may help build efficient, yet cheap to train models.\par
	
	Understanding the response of individual neurons in parts of brain that are highly believed to be responsible for visual object recognition, is quite challenging and hard to predict \cite{DICARLO2012415,brincat2004underlying,yamane2008neural}. Nevertheless, it is known that they are activated by a set of complex visual features and, thus they cannot be narrowly tuned as detectors for specific objects \cite{DICARLO2012415,rust2010selectivity,Desimone2051}. Therefore, single neurons may not be acting as sparse object detectors, but, rather as elements of a group that as a whole, provides object recognition \cite{DICARLO2012415,rust2010selectivity}. However, they are often able to maintain preferences among objects like changes in size and shape which is called neural tolerance \cite{DICARLO2012415,brincat2004underlying}. This suggests that single neurons together decode the object space and identity including position, size, context, and other variables to achieve powerful representation and avoid binding this information at successive stages \cite{DICARLO2012415,dicarlo2007untangling,edelman1999representation,koch2000role}. As an analogy of how brain single neurons behave in visual recognition tasks, artificial neural networks can make use of it as a design concept in developing  populations of single neuron or small-sized networks for object recognition.\par
	
	The design concept introduced here, has proven effectiveness for many classifiers and configurations including SVM and neural networks \cite{Lorena2009, GALAR20111761, natoarticle, anandModular}. The reason it is often successful for classification lies in the fact that it is easier to build classifiers to separate two classes rather than for multiple classes \cite{GALAR20111761}. Usually, the approach is referred to as decomposition \cite{GALAR20111761}, binarization \cite{GALAR20111761}, or modularization \cite{anandModular} and has been addressed in different ways. The first used decomposition strategy is called "one-vs-one" (OVO) and it divides the problem into as many binary classification problems as the number of unique combinations between each two classes \cite{GALAR20111761, natoarticle}. The second strategy is called "one-vs-all" (OVA) and it trains a classifier to discriminate each class from all other classes \cite{GALAR20111761, natoarticle}.\par
	
	Modularization in neural networks has been introduced by \cite{natoarticle} through attempting to separate each class from all other classes and then subsequently pairwise separate each class using subnetworks. Furthermore, \cite{anandModular} used K networks to reduce a K-class problem into a set of K two class problems, however they focused more on the back-propagation algorithm and enhancing its convergence speed. In this work, we used an OVA approach as in \cite{anandModular} and we focused more about the size of modular networks used for each binary sub-classification problem and the performance of the network.\par
	
	This paper addresses the feasibility of neuron population based design approach through answering three main questions: 1. How does the model performance change with the size of hidden layer in the used network? 2. Is it possible that single neuron or small-sized networks can achieve good performance with the binary sub-classification problems of high dimensional multi-class data? 3. Can binary classification networks with low number of hidden neurons (especially single neuron networks) be utilized to perform multi-class recognition? and what is the efficiency of these systems compared to multi-class dense systems? To answer these questions, we experimented multiple architectures along with different levels of abstraction in recognition tasks for three datasets.\par
		
	\section{Methods}
	\label{S:Methods}
	
	\begin{figure*}[!h]
		\centering
		\includegraphics[width=\textwidth]{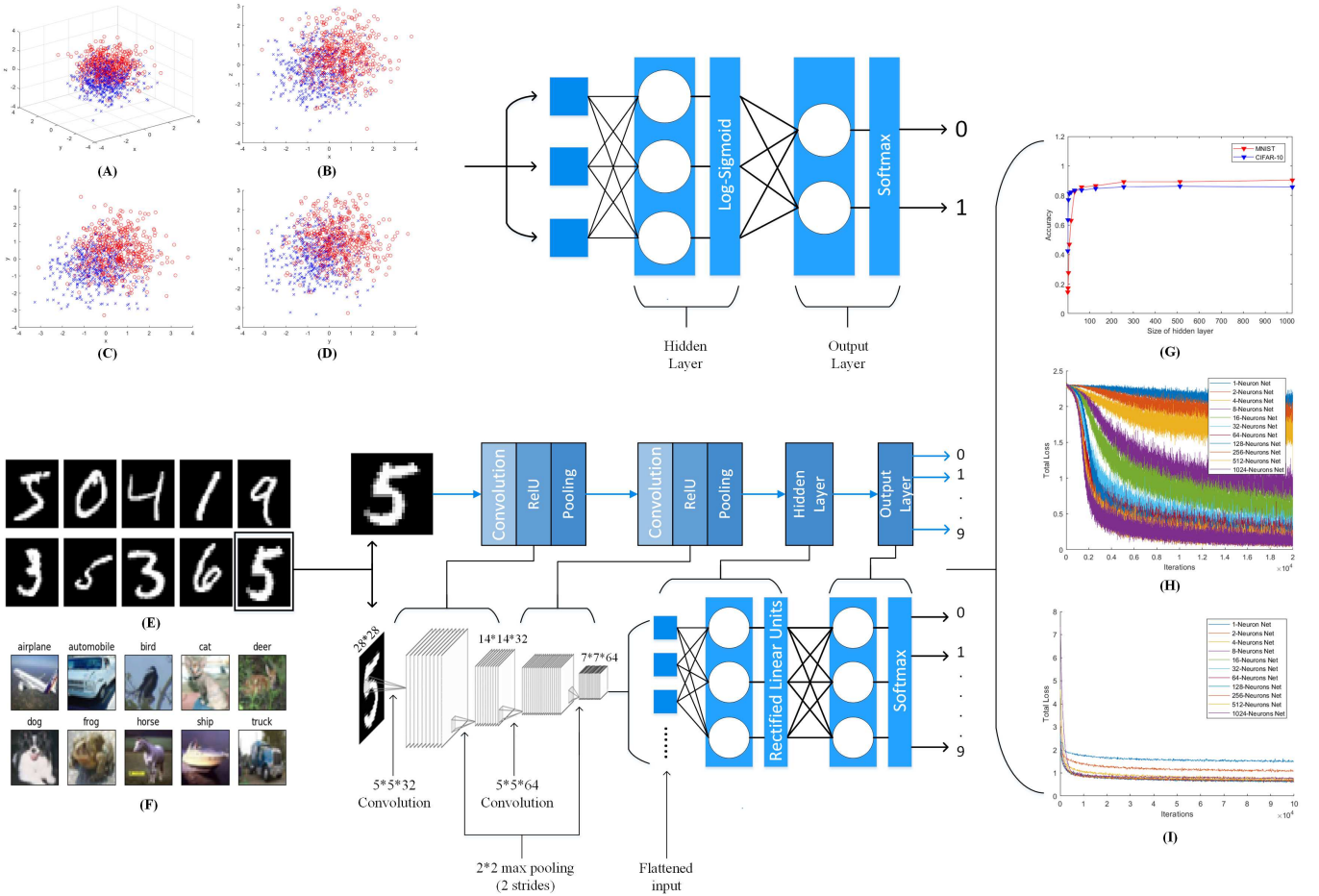}
		\caption{Experimental setup overview. (A) Scatter view of random data drawn from $\mathcal{N}(0,1)$. (B) X-Z projection of the data. (C) X-Y projection of the data. (D) Y-Z projection of the data. (E) Sample of the MNIST dataset. (F) Sample of the CIFAR-10 dataset. (G) Classification accuracy of networks with different hidden layer sizes for MNIST and CIFAR-10 data. (H) Training loss of MNIST dataset for different hidden layer sizes.(I) Training loss of CIFAR-10 dataset for different hidden layer sizes.}
		\label{fig:bdall1}
	\end{figure*}
	
	\subsection{Datasets}
	\label{SS:Datasets}
	To conduct experimental results about the main objectives of this work, we used three main datasets for all trials. The first dataset is composed of 3 dimensional data points drawn from a normal distribution with different values of mean and standard deviation. Particularly, we used a combination of a single value for mean (0) and 3 values for standard deviation (0.1, 0.5, 1). Multiple numbers of data points were used as well for each set, which gives a total of nine sets (3 sets of different lengths drawn from each distribution). Two categories were created inside the generated sets through positively and negatively biasing samples of the data with the same amount. Fig. \ref{fig:bdall1}(a-d) shows a sample of the generated sets with a mean of 0 and standard deviation of 1. The categories for this set were created by adding $0.5$ to half of the set and subtracting the same value from the other half.\par
	
	The second dataset is the famous MNIST \cite{lecunCNN}, widely used for handwritten digit recognition benchmarking. MNIST consists of two sets of images, the first set is 60,000 examples for training and a test set of 10,000 examples. All digits are size-normalized and centered in $28\times 28$ gray scale images (Fig. \ref{fig:bdall1}(e)). MNIST dataset doesn't require formatting or preprocessing which makes it optimal for testing learning techniques and new recognition algorithms \cite{lecunCNN}. The third dataset is the CIFAR-10 dataset \cite{Krizhevsky09} which consists of 60,000 color images each of $32\times 32$ pixels in $10$ classes (Fig. \ref{fig:bdall1}(f)). The dataset is divided into two sets, one for training (50,000 images) and another for testing (10,000 images).\par
	
	\subsection{CNN Design}
	\label{SS:CNNDesign}
	The main architecture of CNN used for all multi-class recognition experiments performed in this analysis, included two convolutional layers, two max pooling layers (one after each convolutional layer), a single hidden dense layer (ReLU activated), and an output layer with a number of units equal to the number of classes in the target dataset. The first convolutional layer applies 32 $5\times 5$ filters with ReLU activation followed by a $2\times 2$ filter max pooling with strides of 2 (non overlapping pooled regions). The second convolutional layer applies 64 $5\times 5$ filters with ReLU activation followed by a $2\times 2$ filter max pooling with strides of 2. For the hidden layer, we used $1, 2, 4, 8, 16,\dots, 1024$ units for the different experiments performed. All experiments developed for this analysis were implemented using Tensorflow\textsuperscript{TM} and Matlab; and tested over an NVIDIA Tesla M40 GPU. \par
	
	\subsection{Effect of Layer Size}
	\label{SS:elayersize}
	The number of hidden neurons used in different network architectures is assessed here through running multiple trials over the 3 datasets mentioned before. First, we trained three feedforward networks for a simple binary classification task over all of the randomly generated sets. Each set was divided into 80\% for training and 20\% for testing of its length. The three networks are of one hidden layer but one with a single neuron, the second with 10 neurons, and the last one with 100 neurons. The average of results was taken over multiple iterations (100 to 1000) to avoid randomness of the experiment \cite{BOYLE1977323}.\par
	
	Second, to see the effect of the number of hidden neurons on more complex recognition problems, we trained a convolutional neural network (CNN) for MNIST handwritten digit recognition and another for CIFAR-10 content recognition. Both networks have two convolutional layers, two pooling layers (one after each convolutional layer), and a single hidden dense layer (ReLU activated). The first convolutional layer applies 32 $5\times 5$ filters with ReLU activation followed by a $2\times 2$ filter max pooling with strides of 2 (non overlapping pooled regions). The second convolutional layer applies 64 $5\times 5$ filters with ReLU activation followed by a $2\times 2$ filter max pooling with strides of 2. For the hidden layer, we used $1, 2, 4, 8, 16,\dots, 1024$ in order to examine the effect of layer size on the recognition task in both datasets.\par
	
	\subsection{Effect of Tearing down the Recognition Problem}
	\label{SS:etdown}
	As mentioned before, dense and deep networks may be needed to model complex recognition problems. However, this will come with a cost, overfitting and expensive training. So, here we test the ability of tearing down multi-class recognition problems into a series of binary classifiers. MNIST and CIFAR-10 datasets were used also for this experiment but with converting each of them into 10 different datasets from the labels point of view. In other words, the labels of each dataset were altered 10 times to classify only one category out of the 10 categories against all the others. For MNIST dataset we have 10 classification problems, the first is to classify the $0$ digit against $1,\ 2,\ \dots,\ and\ 9$, the second is to classify the $1$ digit against $0,\ 2,\ \dots,\ and\ 9$, and so on. The same also is performed for the CIFAR-10 dataset but considering its different categories. All generated datasets were balanced in terms of classes before being used in training.\par
	
\begin{figure*}[!ht]
		\centering
		\includegraphics[width=\textwidth]{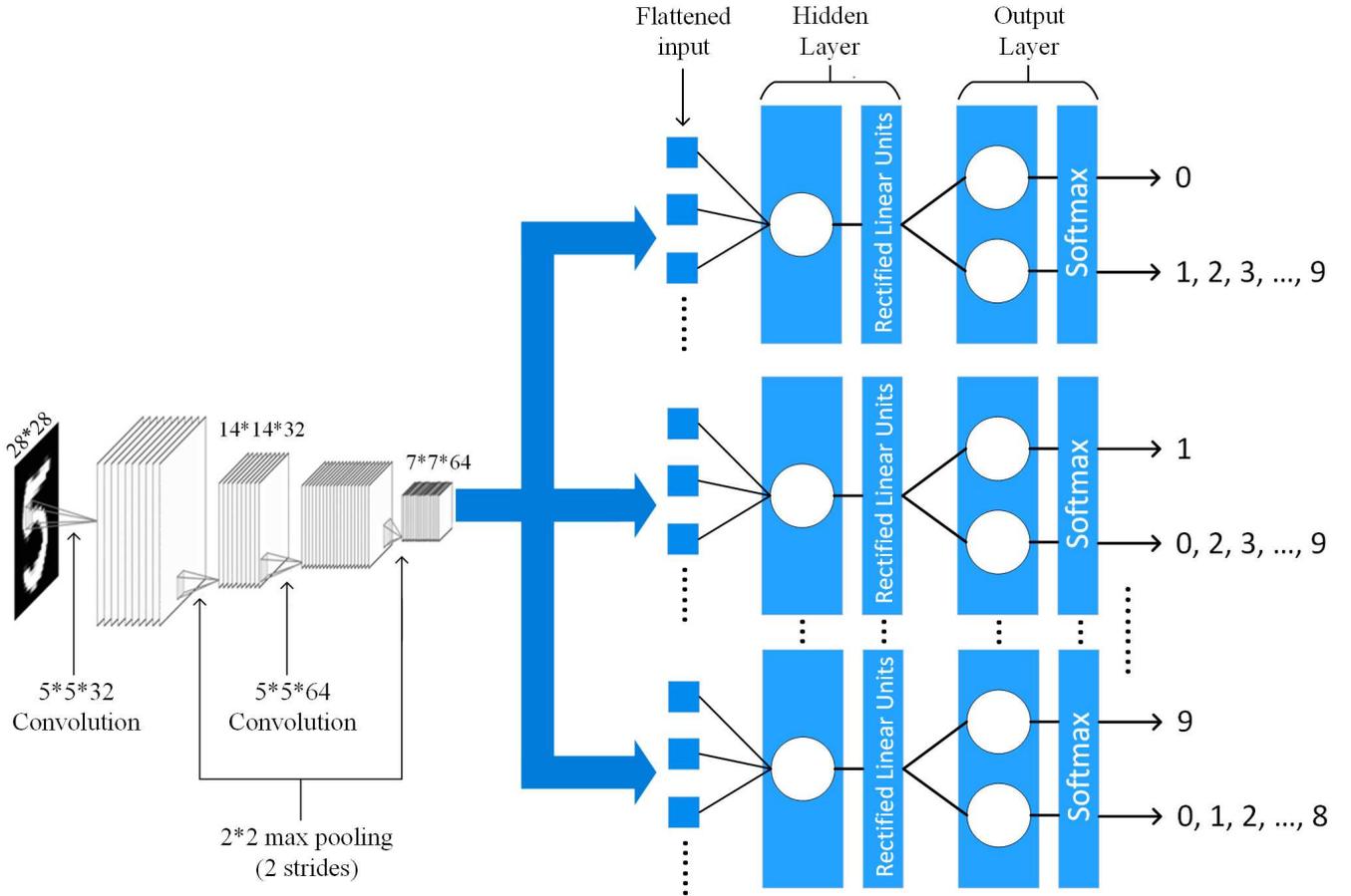}
		\caption{Multi-class recognition using binary classification networks.}
		\label{fig:bdall2}
	\end{figure*}

	We used the same convlutional network mentioned before and tried different sizes for hidden layer too. Ten networks were trained for each dataset and the results were compared with small sized, full classification networks tested with the same datasets. In order to use these parallel networks for 10 classes recognition, the input image is administered to each of the 10 networks and the resultant digit is represented by the network that gives a positive response as in Fig. \ref{fig:bdall2}. Due to the independence of the 10 networks, there might be redundancy in the output (i.e. an image identified more than once throughout the 10 networks). This can be solved through priority encoding the results (take the first positive result and ignore the rest of networks) or consider this as misclassification. For this study, we considered redundancy as misclassification. We saw it as a more convenient way for performance comparison between different architectures.\par
	
	\begin{table}[!h]
		\caption{Performance comparison between single, 10, and 100 neurons networks over random generated data.}
		\label{tab:randdatares}
		\centering
		\tiny{
			\resizebox{\linewidth}{!}{%
				\bgroup
				\def\arraystretch{2}
				\begin{tabular}{c|c|c|c|c|c|c|c|c|c|c}
					\hline
					\hline
					\multicolumn{2}{c|}{Network Size}&\multicolumn{3}{c|}{Single Neuron Net}&\multicolumn{3}{c|}{10-Neurons Net}&\multicolumn{3}{c}{100-Neurons Net}\\
					\hline
					\multicolumn{2}{c|}{Sample Size}&$10^3$&$10^4$&$10^5$&$10^3$&$10^4$&$10^5$&$10^3$&$10^4$&$10^5$\\
					\hline
					&Accuracy&0.693&0.727&0.703&0.771&0.784&0.782&0.753&0.778&0.782\\
					\cline{2-11}
					&Sensitivity&0.695&0.727&0.704&0.775&0.783&0.73&0.758&0.779&0.781\\
					\cline{2-11}
					\multirow{-3}{*}{\begin{turn}{90}
							$\mathcal{N}(0, 0.1)$
					\end{turn}}&Specificity&0.691&0.728&0.702&0.768&0.786&0.782&0.750&0.777&0.782\\
					\hline
					&Accuracy&0.706&0.698&0.700&0.772&0.781&0.782&0.760&0.778&0.782\\
					\cline{2-11}
					&Sensitivity&0.694&0.698&0.701&0.760&0.782&0.783&0.756&0.778&0.781\\
					\cline{2-11}
					\multirow{-3}{*}{\begin{turn}{90}
							$\mathcal{N}(0, 0.5)$
					\end{turn}}&Specificity&0.720&0.698&0.700&0.786&0.782&0.780&0.765&0.778&0.782\\
					\hline
					&Accuracy&0.701&0.6936&0.700&0.776&0.780&0.781&0.756&0.779&0.781\\
					\cline{2-11}
					&Sensitivity&0.699&0.693&0.700&0.775&0.785&0.781&0.767&0.777&0.781\\
					\cline{2-11}
					\multirow{-3}{*}{\begin{turn}{90}
							$\mathcal{N}(0, 1.0)$
					\end{turn}}&Specificity&0.703&0.693&0.699&0.778&0.775&0.781&0.747&0.780&0.782\\
					\hline
				\end{tabular}
				\egroup
		}}
	\end{table}
	
	\section{Results}
	\label{S:Results}
	Changing the hidden layer size in a neural network will definitely affect the performance of the network, but, the question is whether the change in performance is worth it or not. Testing networks of different hidden layer sizes using the random generated data gave the results shown in Table \ref{tab:randdatares}. On the other hand, testing over a wider range of hidden layer sizes for multi-class complex data showed that increasing the number of hidden neurons is not quite effective after a certain point. The accuracy of 10 classes recognition task for both MNIST and CIFAR-10 remains nearly constant after a hidden layer size of 128 neurons as shown in Fig. \ref{fig:bdall1}(g). Probably this size will be different from a dataset to another and even from a recognition task to another, however, this shows that the same performance can be achieved with way smaller networks. No changes in the total loss pattern were observed as well after this layer size as shown in Fig. \ref{fig:bdall1}(h) and (i). In addition to that, small sized networks were found to suffer from high loss despite of faster convergence which can be clearly seen in CIFAR-10 total loss in Fig. \ref{fig:bdall1}(i).\par
	
		\begin{table}[!h]
		\caption{Performance of tearing down multi-class problems into binary classification with the use os small sized hidden layers.}
		\label{tab:binarymnistcifarres}
		\centering
		\bgroup
		\def\arraystretch{1.5}
		\begin{tabular}{c|l|c|l|c}
			\hline
			\hline
			Dataset&\multicolumn{2}{c|}{MNIST/Single Neuron Net}&\multicolumn{2}{c}{CIFAR-10/128 Neurons Net}\\
			\hline
			&Zero against all&0.982&Airplane against all&0.959\\
			\cline{2-5}
			&One against all&0.985&Automobile against all&0.976\\
			\cline{2-5}				
			&Two against all&0.978&Bird against all&0.952\\
			\cline{2-5}
			&Three against all&0.960&Cat against all&0.948\\
			\cline{2-5}
			&Four against all&0.997&Deer against all&0.955\\
			\cline{2-5}
			&Five against all&0.985&Dog against all&0.944\\
			\cline{2-5}
			&Six against all&0.985&Frog against all&0.960\\
			\cline{2-5}
			&Seven against all&0.975&Horse against all&0.963\\
			\cline{2-5}
			&Eight against all&0.956&Ship against all&0.977\\
			\cline{2-5}
			\multirow{-10}{*}{\begin{turn}{90}
					Accuracy
			\end{turn}}&Nine against all&0.932&Truck against all&0.961\\
			\hline
		\end{tabular}
		\egroup
	\end{table}
	
	Since over-reduction of network size alone, shows poor performance in multi-class recognition problems as it appears in Fig. \ref{fig:bdall1}(g), simplifying the recognition problem might be the solution. Given the performance of single neuron networks for binary classification in Table \ref{tab:randdatares}, we test the ability of tearing down multi-class recognition problems into a set of binary classifiers using small sized networks as in Fig. \ref{fig:bdall2}. Table \ref{tab:binarymnistcifarres} shows a sample of the results for tearing down the MNIST and CIFAR-10 datasets into 10 different binary classification problems. It seems that using a small sized hidden layer along with a binary problem gives better results than the best performance achieved using high density layers used with multi-class recognition problems considering all classes. For MNIST problem, good results were achieved using a single neuron in hidden layer. On the other hand, CIFAR-10 classification accuracy did not jump over 80\% before using 64 neurons in the hidden layer.  \par
	
	The 10 networks trained for binary classification problems for MNIST were combined together to form a 10-classes recognizer as in Fig. \ref{fig:bdall2}. The new system correctly identified 84.21\% of the test data, produced multiple classification results for 12.79\% of the test data (maximum of two positive results per image appeared), and the rest of test data (3\%) were misclassified. For CIFAR-10 dataset, to achieve the same good results as in MNIST experiment, we used 10 networks with at least 128 neurons per hidden layer for each, which will be more complex and ineffective than using a full classification network with the same number of neurons in hidden layer.\par
	
	\section{Discussion}
	\label{S:Discussion}
	The experimental results of this work, showed that neural networks of dense hidden layers might not be of a great help to achieve the desired modeling of the recognition/classification task. For a binary classification task, increasing the hidden layer size did not add much to all the aspects of system performance as shown in Table \ref{tab:randdatares}. Even in complex multi-class recognition tasks like digits and objects identification, the performance becomes nearly the same after a certain layer size. This characterizing layer size will probably depend on the level of abstraction of the assessed problem, however, we can clearly see that good performance can be achieved by using fairly sparse networks.\par
	
	An acceptable performance can be easily achieved in simple classification tasks using small sized networks and becomes harder in high dimensional tasks. This can be noticed through the differences between training networks for binary classification task and multi-class tasks (Table \ref{tab:randdatares} and Fig. \ref{fig:bdall1}(g)). The classification accuracy is nearly stable at low network density for the binary random data, but needs more hidden neurons to get to the same stable performance for MNIST and CIFAR-10 tasks. This suggested tearing down multi-class recognition tasks to multiple binary classification tasks for which, fast convergence, more simple architectures, and acceptable performance can be reached easier.\par
	
	Using the binary classification scheme for both MNIST and CIFAR-10 datasets, gave superior performance even with using a single neuron hidden layer. Compared to the highest accuracy achieved for 10-classes recognition, the binary scheme achieved higher classification accuracy for all components. This proves our claim about using populations of binary systems to represent higher dimensional datasets for a better performance and cheap training.\par
	
	The high accuracy achieved in binary classification networks pushed toward building multi-class recognition based on these networks. Parallel sparse hidden layer, binary networks with the same number as desired classes, were used to build multi-class classifiers with a higher accuracy than a single multi-class network with the same number of hidden neurons used for the same task. A single network with 16 neurons in the hidden layer got a classification accuracy of 82\% (Fig. \ref{fig:bdall1}(g)) for MNIST dataset while 10 single neuron binary networks achieved around 84\% accuracy. The low value of contradicting results from each of the 10 combined networks comes from the fact that the accuracy of each single network is very high that there is a very low chance that an image will get identified in more than one network. \par
	
	The experiments performed in this study, showed that robust results can be achieved using small number of hidden neurons and these results were confirmed by multiple trials on different datasets. The experiments showed also that using neuron populations in artificial object recognition can achieve a similar performance pattern as anticipated from the biological models. However, it must be taken in consideration that we tested the design concept for multi-object recognition and not for the same object attributes like shape, size, color, and rotation. This can be an indicator that increasing the level of representation may be in favor of the recognition performance. In other words, adding a neuron to the population to tolerate more visual changes of objects may help achieve better recognition performance between objects. \par
	
	To conclude, we assessed the effect of reducing the hidden layer size in neural networks on the performance of different recognition tasks including binary random data and 10-classes recognition problems. The results showed that high performance can be achieved using fairly small-sized networks from the number of hidden neurons prospective. Moreover, we assessed the use of a population of small sized binary networks in building multi-class recognition systems and it showed superior results compared to multi-class systems with the same hidden layer size. There is more to build over these preliminary findings, therefor, we intend to test more visual aspects in object recognition with neuron populations for better generalization of the proposed design concept.\par
	
	\bibliographystyle{IEEEtran}
	\bibliography{MyBibliography}
\end{document}